\documentclass{article}

\usepackage{arxiv}

\usepackage[utf8]{inputenc} 
\usepackage[T1]{fontenc}    
\usepackage{hyperref}       
\usepackage{url}            
\usepackage{booktabs}       
\usepackage{amsfonts}       
\usepackage{nicefrac}       
\usepackage{microtype}      
\usepackage{lipsum}
\usepackage{graphicx}

\title{Synthetic Lunar Terrain: A Multimodal Open Dataset for Training and Evaluating Neuromorphic Vision Algorithms}

\author{
 Marcus Märtens \\
  Australian Institute for Machine Learning\\
  University of Adelaide\\
  \texttt{marcus.martens@adelaide.edu.au} \\
  \And
 Kevin Farries \\
  Andy Thomas Centre for Space Resources\\
  University of Adelaide\\
  \texttt{kevin.farries@adelaide.edu.au} \\
  \And
 John Culton \\
  Andy Thomas Centre for Space Resources\\
  University of Adelaide\\
  \texttt{john.culton@adelaide.edu.au} \\
  \And
 Tat-Jun Chin\\
  Australian Institute for Machine Learning\\
  University of Adelaide\\
  \texttt{tat-jun.chin@adelaide.edu.au} \\
}

\begin{document}
\maketitle
\begin{abstract}
Synthetic Lunar Terrain (SLT) is an open dataset collected from an analogue test site for lunar missions, featuring synthetic craters in a high-contrast lighting setup. It includes several side-by-side captures from event-based and conventional RGB cameras, supplemented with a high-resolution 3D laser scan for depth estimation. The event-stream recorded from the neuromorphic vision sensor of the event-based camera is of particular interest as this emerging technology provides several unique advantages, such as high data rates, low energy consumption and resilience towards scenes of high dynamic range. SLT provides a solid foundation to analyse the limits of RGB-cameras and potential advantages or synergies in utilizing neuromorphic visions with the goal of enabling and improving lunar specific applications like rover navigation, landing in cratered environments or similar.
\end{abstract}


\section{Introduction}
\label{sec:introduction}

The Moon holds a vast significance for mankind, with the number of dedicated missions seeing a recent resurgence~\cite{LunarExploration}. Both, institutional actors like (inter)national space agencies and commercial entities like SpaceX, Blue Origin, Astrobotic or iSpace have sent or are planning to send their orbiters or landers to several regions of interest. While there is high scientific and economic potential in lunar operations, there are also risks involved related to the numerous challenges of this extreme and under-explored environment.

Areas of particular interest on the Moon are the south polar regions with some of its oldest terrains~\cite{Krasilnikov2023Geologic} and unique lighting conditions~\cite{Mazarico2010Illumination}. The oblique solar incidence angle, the specific reflectance properties of the lunar regolith and the general absence of an atmosphere contribute to an extreme-contrast situation, characterized by partially illuminated craters, elongated drop-shadows and areas of permanent illumination or permanent darkness. Any vehicle, such as a lunar rover or a lunar lander, needs to be equipped with adequate perception capabilities, both in hardware and software, to safely navigate those areas. 

\begin{figure}
  \centering
    \includegraphics[scale=0.2]{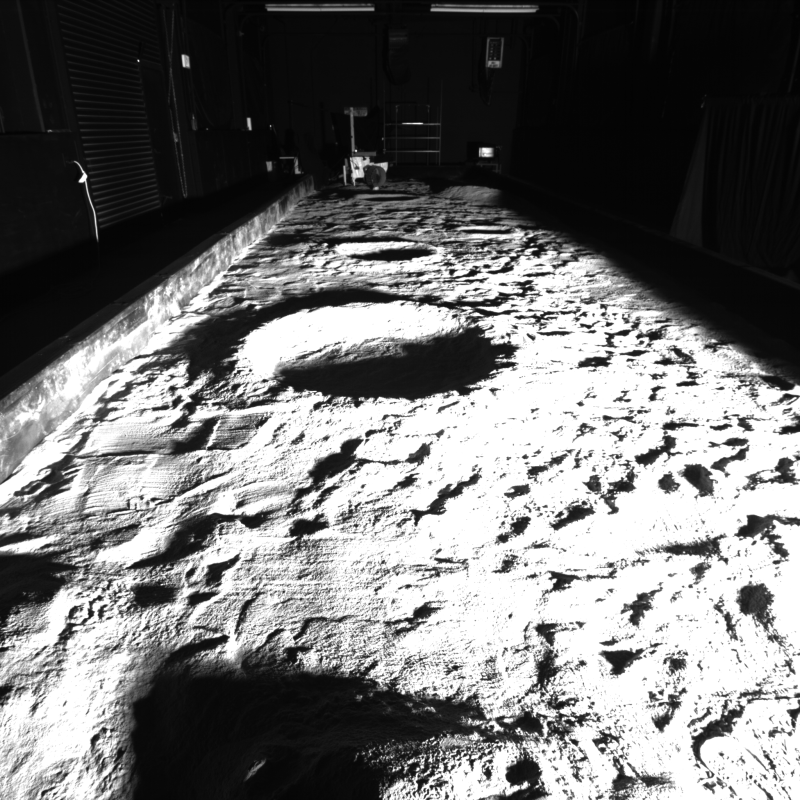}  
    \includegraphics[scale=0.255]{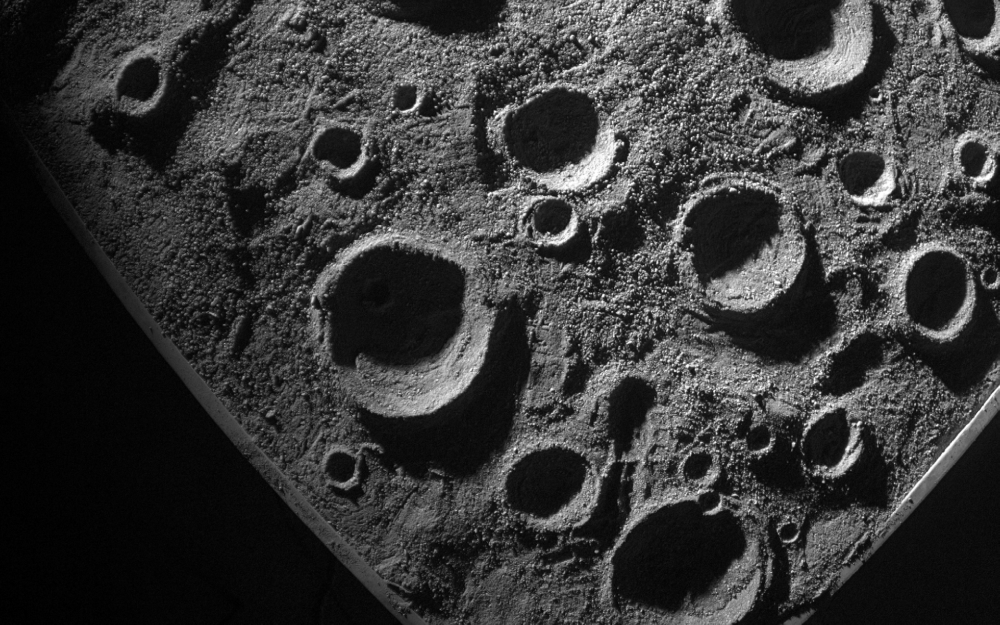}  
  \caption{Comparison of one example from POLAR~\protect\cite{Hansen2023polar} (left) with RGB-modality of SLT~\protect\cite{Martens2024synthetic} (right).}
  \label{fig:polar_comparison}
\end{figure}

On the hardware side, neuromorphic vision sensors are an emerging technology that has been proposed for space applications in general~\cite{Izzo2022Neuromorphic}. A neuromorphic pixel will generate a binary event if its change in brightness crosses a sensitivity threshold. Moving away from absolute brightness to brightness changes makes neuromorphic sensors exceptionally robust for high dynamic range (HDR) situations, with a fast data-rate and a comparatively low power consumption as added benefits.

On the software side, machine learning approaches have shown tremendous potential in the field of computer vision~\cite{Mahadevkar2022Review}. The most common approach, deep supervised learning, requires a large, annotated dataset to be ingested to calibrate the parameters of a large artificial neural network. Once trained at examples on a particular task, these networks exhibit the capability to generalize to unknown situations. The final performance depends largely on the quantity and quality of their training data. For a task like navigating the lunar landscape, the best case would be to train the neural network on annotated footage directly taken from the Moon. Undoubtedly, this creates a ``chicken-and-egg''-situation, as such type of data is currently extremely scarce and would arguably require some degree of navigational capabilities to be obtained to begin with.

A possible way to circumvent this dilemma is the utilization of computer graphics to generate digital images for training, mimicking the general characteristics one would expect to encounter on the Moon~\cite{Pessia2019Artificial}. Although computer graphics allow for the generation of vast quantities of synthetic data, the gap in realism can be significant, since important details such as high-fidelity lighting physics are challenging to reproduce. In order to narrow the gap, analogue reconstructions that emulate lunar environmental conditions directly provide an alternative way of generating richer and, arguably, more realistic data. The POLAR Traverse Dataset created by NASA is a noteworthy example~\cite{Hansen2023polar}. This dataset provides stereo pairs of camera images taken from a test bed filled with regolith simulant under varying lighting conditions and exposures. A 3D scan of the synthetic landscape is provided as ground truth for its geometry.

Our contribution is a new open dataset, called Synthetic Lunar Terrain (SLT), providing a multi-modal capture of an analogue test bed with a large amount of craters, created at the  Exterres Analogue Facility. Similar to POLAR, SLT provides a high-resolution 3D scan of its geometry for accurate depth estimation. In contrast to POLAR, SLT includes the event-stream of a neuromorphic vision sensor as an additional modality. Moreover, SLT was recorded using a 9kW metal-halide lamp as source of illumination, providing an extreme lighting situation with challenges in contrast. With the publication of our dataset, we aspire to enable research into computer vision with a focus on neuromorphic sensing in particular. Table~\ref{tab:table} shows a short comparison between some of the characteristics of the POLAR and the SLT dataset.

This work documents our preliminary investigations, the data collection process and the details of the initial release of SLT. It is structured as follows: Section~\ref{sec:data_collection} describes the data collection process from the lunar surface in detail. Section~\ref{sec:data_post-processing} includes a description of the three modalities of SLT, the organization of the data and the post-processing steps that have been undertaken. Section~\ref{sec:analysis} highlights several observations and characteristics of the dataset, including its limitations. Finally, Section~\ref{sec:conclusions} concludes with an outlook about possible applications and ways forward supported by the SLT data.

\begin{table}
 \centering 
 \def\arraystretch{1.2}
 \begin{tabular}{|l|c|c|}
 \hline 
 ~ & \textbf{POLAR~\cite{Hansen2023polar} (NASA)} & \textbf{SLT~\cite{Martens2024synthetic} (ours)} \\ 
 \hline 
 Modalities & Stereovision RGB, 3D point cloud & Monocular RGB, event-streams, 3D point cloud \\ 
 \hline 
 Viewing angles & 14-35\textdegree & 90\textdegree ~(top-down) \\ 
 \hline 
 Camera positions & 
\begin{tabular}{@{}c@{}}Stationary \\ (with approximated camera poses)\end{tabular} 
 & 
\begin{tabular}{@{}c@{}}Sweeping motion during recording \\ (no camera poses)\end{tabular} \\ 
 \hline 
 Exposures & 15 & 1 \\ 
 \hline 
 Traverses & 24 & 42 \\ 
 \hline 
 Terrain material & Modified LHS-1 regolith simulant & \begin{tabular}{@{}c@{}} 50\% GGBS mixed with \\  50\% Kanmantoo crusher dust\end{tabular} \\ 
 \hline 
 Terrain features & 
\begin{tabular}{@{}c@{}}Focus on representative landscape, \\  4 large craters\end{tabular} 
 & 130 craters of varying sizes with overlaps \\ 
 \hline 
 3D scan resolution & 8mm & <2mm \\ 
 \hline 
 Size total dataset & 13.4GB & 127.6GB \\ 
 \hline 
 \end{tabular} \\
 \vspace*{0.5em}
 \caption{Comparison between selected dataset characteristics between POLAR and SLT.}
 \label{tab:table}

\end{table}

\section{Data Collection}
\label{sec:data_collection}

Our synthetic lunar surface was created  at the Exterres Lunar Analogue Facility. On an area of 3.6m x 4.8m, a scene containing varying elevations and surface features was sculpted using custom-made regolith simulant as material. Our simulant was produced by mixing 50\% ground granulated blast furnace slag with 50\% Kanmantoo Crusher dust, achieving an albedo of 0.14, comparable to numbers reported on lunar highland material. Over one hundred craters of varying sizes and depths were created by slowly rotating a T-shaped template, scraping the material out and forming rims, with occasional overlaps, including small craters placed in larger ones.

The whole scene was walled-in with black-out fabric to reduce the impact of unwanted external light sources and reflections. The designated source of light for the scene was coming through a small, rectangular opening in one of the surrounding walls, 2.4m from the testbed. On the other side of this opening, an additional 3.3m away, we operated a 9kW metal-halide lamp, which provided a constant source of high intensity lighting at an oblique angle of about 20\textdegree, illuminating most of the scene.

\begin{figure}
  \centering
    \includegraphics[scale=0.12]{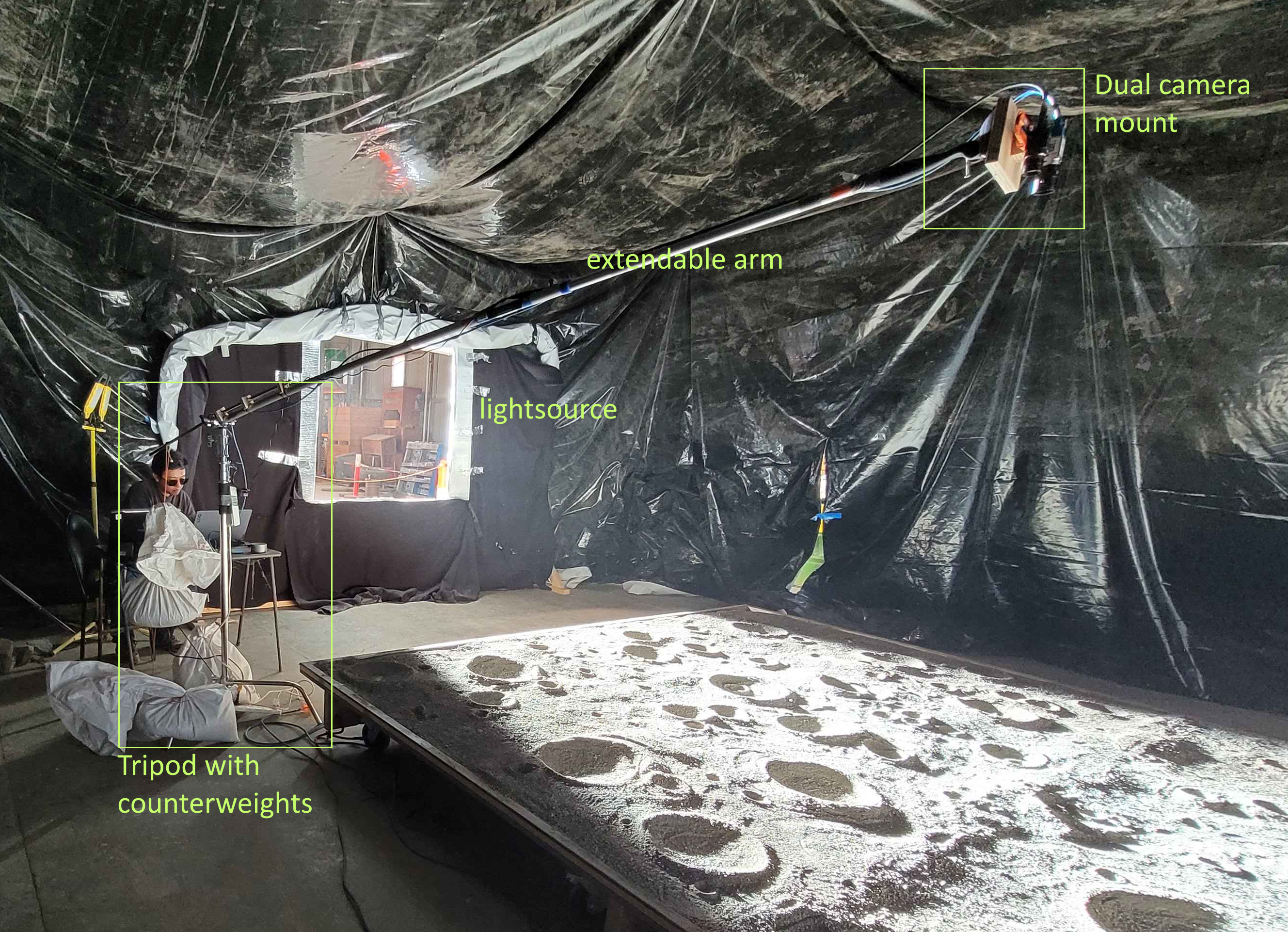}  
  \caption{Lunar Terrain at Exterres Lab, experimental setup.}
  \label{fig:exp_setup}
\end{figure}

The first two modalities of the data were recorded using a conventional monocular RGB camera (Basler a2A1920-160ucPRO with Sony IMX392 sensor) and an event-based camera (Gen4 Prophesee with Prophesee-Sony IMX636 neuromorphic sensor). Given the entirely static scene and the stable, unchanging illumination, simply pointing the event-camera at the scene would have only resulted in noise, as the neuromorphic vision sensor requires brightness changes to generate any events. Consequently, the camera needed to be moved across the scene to generate a meaningful signal. For this purpose, the RGB and the event-camera were both mounted side-by-side, downward-pointing at the end of an extendable rod, attached and counterweighted on a tripod, functioning as a jib-arm. Placing the tripod at varying positions, the jib-arm was manually moved roughly 2.8m above the surface in a slow and continuous sweep. Given the lighting setup and length of the jib-arm, it was possible to record footage without introducing any shadows from the mount itself. Figure~\ref{fig:exp_setup} shows a photo of the setup.

In total, RGB image sequences and event-data streams were recorded from 21 different positions with both, clockwise and anti-clockwise sweeps. Both cameras were focused and calibrated before taking recordings. The specific settings of the cameras remained unchanged during most of the data collection phase. However, for a few locations, we recorded additional data with modified biases and exposure settings or by placing an obstacle into the light-path to study the corresponding impact on the image and event-stream generation. 
The third modality was recorded by placing a FARO Focus S70 3D scanner at the four corners of the testbed, creating a 3D scan of the surface. All raw data of SLT was collected within a single day and care was taken to avoid disturbing the scene. To the best of our knowledge, all modalities are consistent with each other, displaying the same, unchanged scene.

\section{Data Post-processing}
\label{sec:data_post-processing}

\begin{figure}
  \centering
    \includegraphics[scale=0.12]{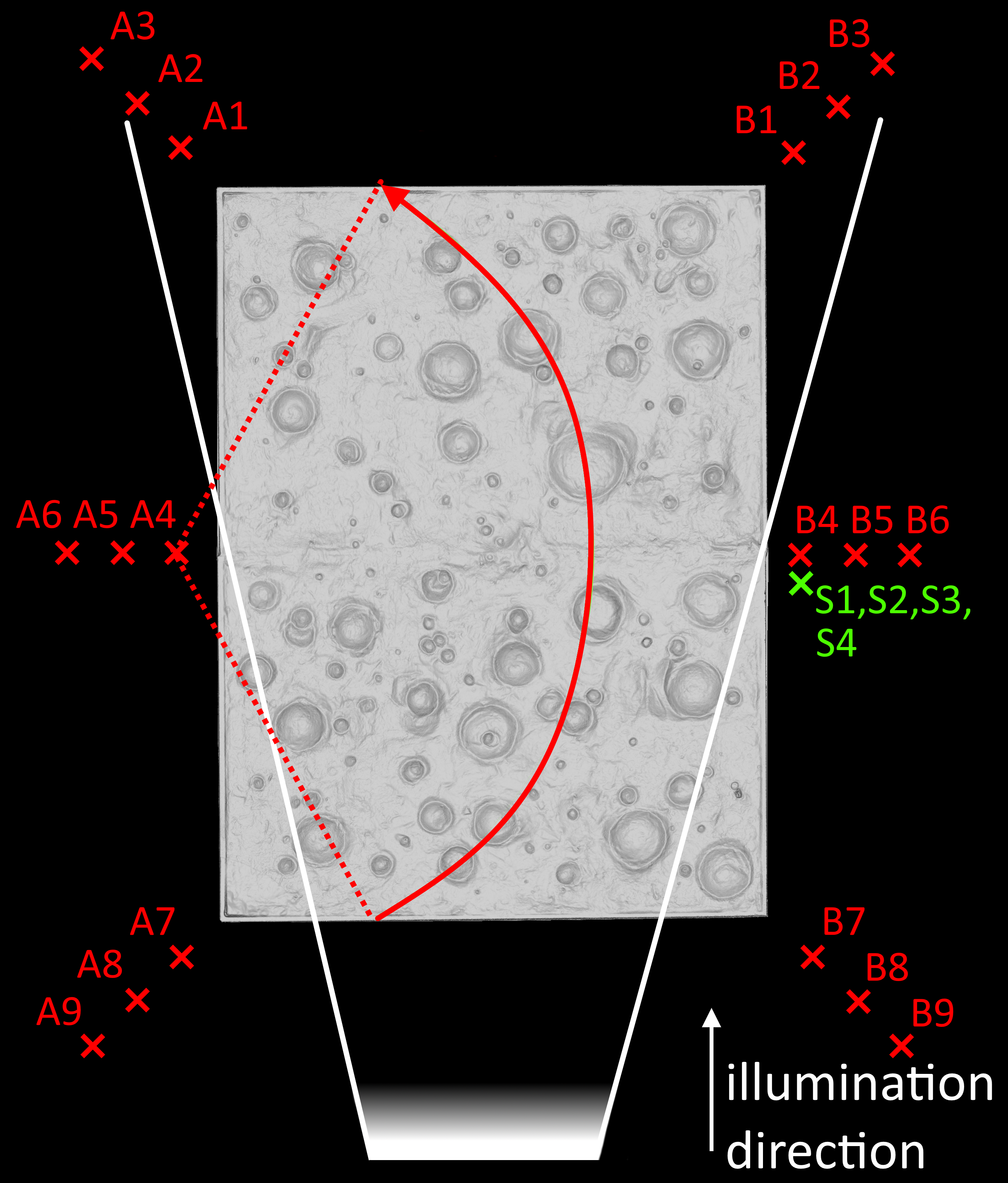}  
  \caption{Example of anti-clockwise recording sweep over the terrain from position A4. All positions from which the jib-arm was moved are shown.}
  \label{fig:positionmap}
\end{figure}

The recorded data from the RGB and the event-based camera was organised according to the position of the mount (A1-A9 for left side, B1-B9 for right side and S1-S4 for modified exposure settings) and the directionality of the sweep (CW for ``clockwise'' and ACW for ``anti-clockwise'' motion). Figure~\ref{fig:positionmap} shows the different positions and an example sweep from the A4 position.

The data of the RGB camera is provided as consecutive image sequences in .tif-format, applying LZW compression at 1920x1200px 8bit grayscale. The length of the image sequences varies, but is usually somewhere between 300 to 700 images per traverse. Duplicate images, recorded especially before the start of the camera motion, and images that show more than ~50\% content not belonging to the terrain (i.e. the floor of the facility) were removed.
The data from the event-camera was recorded at 1280x720px in the binary EVT 3.0 format~\cite{Prophesee}, a format that maximizes the compactness of the event information. For convenience, the dataset includes a standalone C++ script that allows the conversion from EVT 3.0 into a .csv-file, containing the following information in plaintext:
\begin{itemize}
\item timestamp of the event
\item x/y-coordinate of the pixel
\item polarity of the event (i.e. changing from a lower to a higher brightness or vice versa)
\end{itemize}
For further visualization purposes, each event-stream has been converted into a .mp4 video file, using an event accumulation window of 33.3333ms. Similarly to the RGB data, redundant or irrelevant footage was removed. In addition to the event-stream data, a bias-file including the most important camera settings (i.e. sensitivity of the neuromorphic sensor) is provided with each event recording.

\begin{figure}
  \centering
    \includegraphics[scale=0.3]{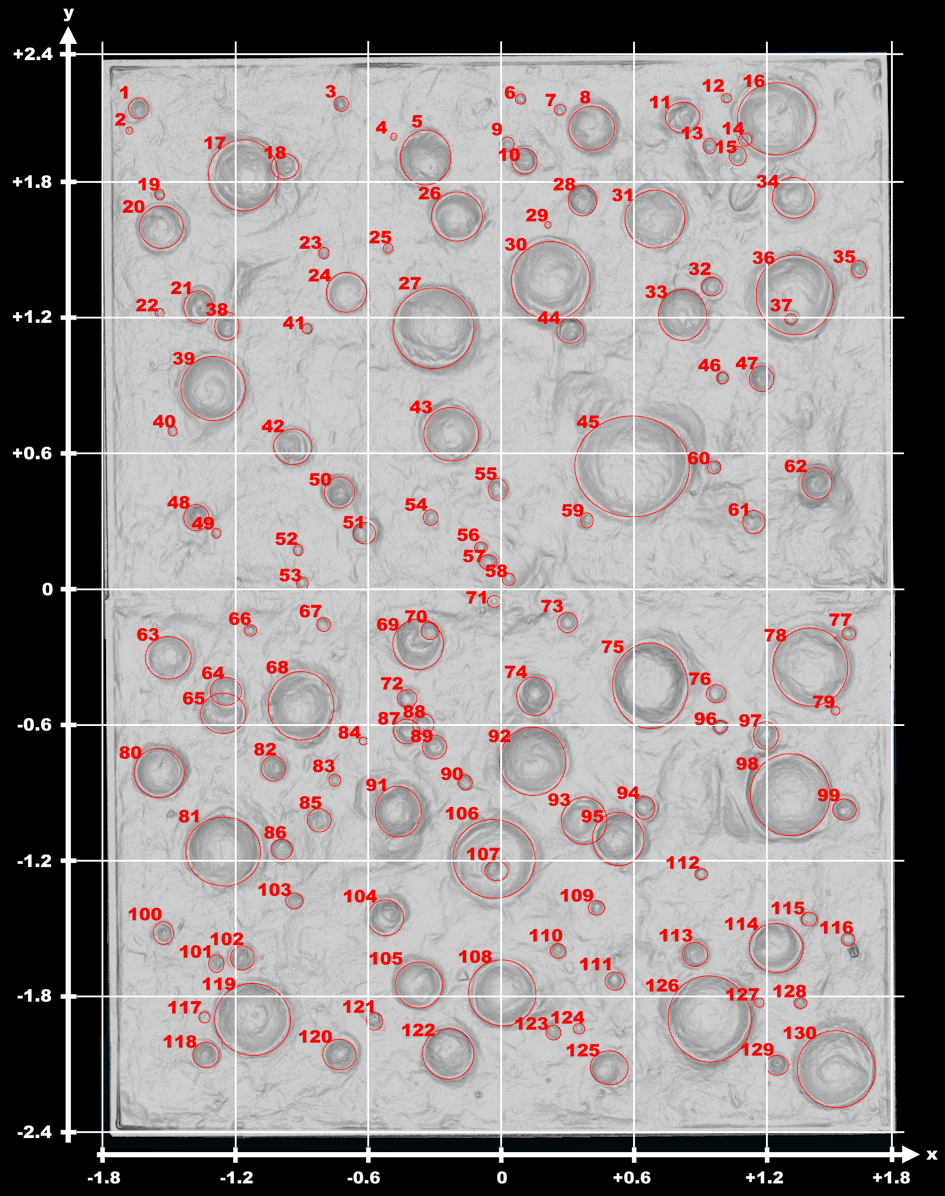}  
  \caption{Map of manually identified crater-like structures within global coordinate system, top-down view.}
  \label{fig:cratermap}
\end{figure}

The data from the 3D scan were registered and merged into a single 3D point cloud. Points not belonging to the terrain were removed manually. Next, the point cloud was decimated to a size of about 6.25M points using Poisson disc sampling, providing an average surface density of 1.862p/mm². Lastly, the coordinates of the point-cloud were transformed, placing the origin of the global coordinate system at the centre of the scene and (roughly) aligning the x-axis with the shorter side of the terrain and the y-axis with its longer side (see Figure~\ref{fig:cratermap}). The z-axis corresponds to a change in elevation and the direction of the light is always towards the positive direction of the y-axis. The final point cloud is provided in the Stanford Polygon File Format (.ply) and as a plaintext ASCII file (.xyz). 
The entirety of the post-processed dataset was released on the open data repository Zenodo under Creative Commons Attribution 4.0 International License~\cite{Martens2024synthetic}. Altogether, the full dataset (zip-compressed) is 127.6GB in size. A smaller 2.2GB sample is available under the same URL for convenience.\footnote{\url{https://zenodo.org/records/13218780}}

\section{Analysis, Observations and Limitations}
\label{sec:analysis}

Based on the 3D point cloud of the terrain, a total of 130 crater-like features were manually identified. A corresponding map of those craters was generated, allowing  their position in x/y-coordinates to be roughly inferred (see Figure~\ref{fig:cratermap}).

\begin{figure}
  \centering
    \includegraphics[scale=0.29]{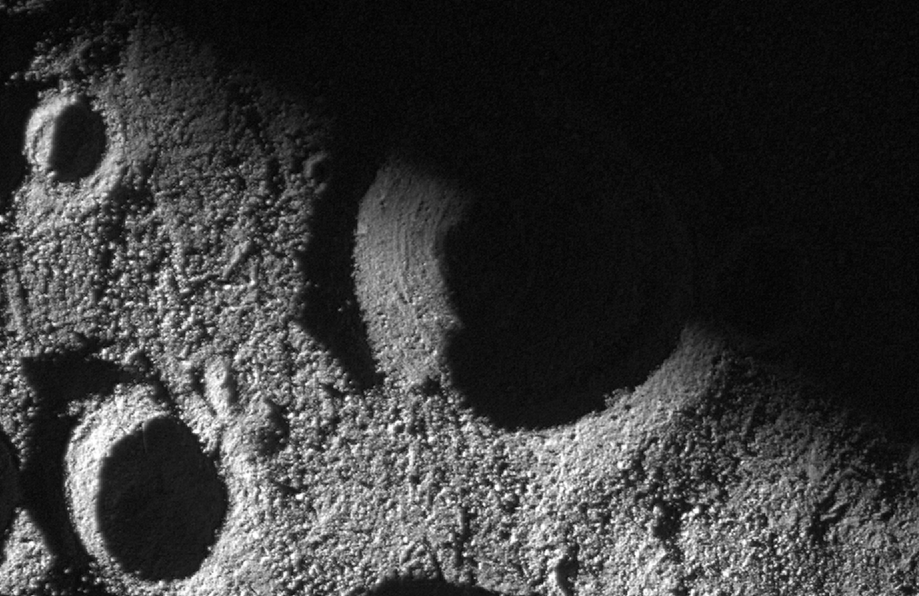}  
    \includegraphics[scale=0.24]{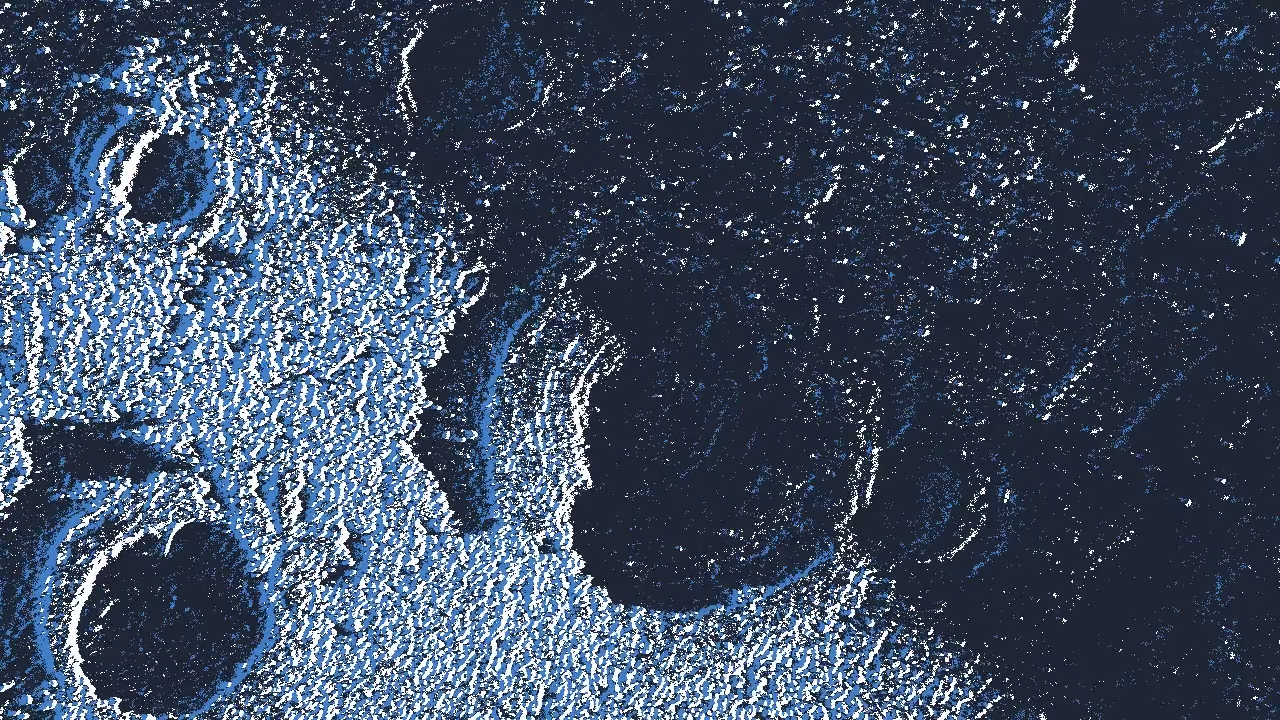}  
  \caption{Example of terminator in RGB-image left (A4\_ACW\_frame296.tif) and event-stream (A4\_ACW.raw) right.}
  \label{fig:event_sample}
\end{figure}

A comparison of the accumulated event-frames with images taken by the RGB-camera confirmed that the latter suffers from over- or underexposure in many situations. This issue becomes even more apparent in the views that contained an obstacle between the terrain and the lighting-source, creating images that are partly under extreme illumination and partly under extreme darkness. This effect is reminiscent of the day/night terminator of the Moon itself. Conventional cameras do not have sufficient dynamic range to accurately resolve features in both, the dark and bright areas, simultaneously. In contrast, although the terminator remains clearly visible in the event-based data as well, the craters appear, despite lower resolution, sharper on both sides of it (see Figure~\ref{fig:event_sample}).

Given the above observations and the richness of the collected data, we are confident that its quality is high enough to facilitate meaningful research, as it is, to the best of our knowledge, the first dataset containing event-streams from synthetic lunar landscapes with conventional images side-by-side. However, SLT is not without shortcomings that need to be considered when working with it.

First and foremost, the sculpted craters were created using a vastly different process and in a different material than the craters on the Moon. Our material was selected to be safe, cost-effective and representative of Moon albedo, but has a larger particle size than true lunar regolith. While there was attention given to create the most important features of a moon-crater including a sharp crater-rim, flat bottom area and almost perfectly circular shape, not all craters in SLT fulfil all three criteria. Moreover, the general size-distribution of the craters was limited by the templates used to create them and is not representative of the actual size-distribution on the moon. Also, the annotations of the craters, as of the initial release of the dataset, are currently only available as an identifier instead of a segmented region.

Although the RGB and the event-camera recordings were done simultaneously, they are not synced against each other and are taken from slightly different perspectives and with different Field-of-Views (FOVs). Moreover, apart from filtering out some garbage images, no post-processing was done to correct for optical artifacts or lens distortions. More precisely, a fish-eye effect is prominently visible in the RGB-images that use a wider FOV. The framerate at which the RGB-camera was taking images was also not constant and depended on the content of the image.

Due to the manual setup of the jib-arm, a perfectly nadir-pointing camera is not guaranteed. Furthermore, the sweeping motion was done by hand and consequently comes with some natural variation in speed and stability. This also means that the clockwise and anti-clockwise sweeps, although technically taken from the same position, might still cover slightly different terrain. Moreover, the motion of the jib-arm could not be guaranteed to be perfectly continuous and suffers from an occasional ``yank'', particularly notable in the event-data.

Given the dustiness of the regolith simulant and the surrounding environment/atmosphere, it was unavoidable to have some quantity of particles in the air above and around the surface, diffracting and reflecting some of the incoming light. Even though all of the interior of the lab was darkened using black-out fabrics, stray light could not be eliminated entirely, which has some impact on the quality of the shadowy areas of some of the craters.

Lastly, mother nature sent an insect to disturb our experiments, which appears in some rare cases as a small blob of fast-moving pixels in the event-stream. While such insects have not been observed on the Moon so far, this ``bug'' nevertheless demonstrates the impressive speed and capabilities of the neuromorphic vision sensor, which is why we decided to keep it in.

\section{Conclusions and Future Directions}
\label{sec:conclusions}

Given the potential of neuromorphic vision for space applications, the open availability of related datasets like SLT are valuable catalysts for scientific discovery and the development of new vision algorithms under the event-based paradigm. Given the multi-modality of SLT, the high-precision ground-truth 3D data and the large quantity of event-recordings, we are optimistic to contribute towards such efforts. Preliminary investigation in phenomena like the artificial terminator give reason to believe that neuromorphic vision sensors can provide unique benefits for the lunar environment specifically. Obvious tasks like crater-detection and segmentation on SLT are already on their way and it is our plan to supplement SLT with additional annotations and studies in the near future. In the long run, some of the lessons learned from creating SLT might enable the generation of new datasets of greater fidelity by specifically addressing some of the design weaknesses and limitations of SLT. For example, future datasets should provide synchronized image taking, potentially sharing the same optics with the cameras mounted on rails or robotic arms for increased controls and accurate poses.

\bibliographystyle{unsrt}

\begin{thebibliography}{1}

\bibitem{Martens2024synthetic}
Marcus Märtens, Kevin Farries, John Culton and Tat-Jun Chin.
\newblock Synthetic Lunar Terrain: A Multimodal Open Dataset for Training and Evaluating Neuromorphic Vision Algorithms [Data set].
\newblock In {\em International Symposium on Artificial Intelligence, Robotics and Automation in Space, i-SAIRAS},2024.
\newblock doi: \url{https://doi.org/10.5281/zenodo.13218780}

\bibitem{Hansen2023polar}
Margaret Hansen, Uland Wong and Terrence Fong.
\newblock Polar Optical Lunar Analog Reconstruction (POLAR) Traverse Dataset.
\newblock {\em NASA Ames Research Center}, 2023.
\newblock \url{https://ti.arc.nasa.gov/dataset/PolarTrav/}

\bibitem{LunarExploration}
\newblock {Lunar Exploration Timeline.}
\newblock {\em NASA},
\newblock \url{https://nssdc.gsfc.nasa.gov/planetary/lunar/lunartimeline.html}

\bibitem{Krasilnikov2023Geologic}
S.S. Krasilnikov, M.A. Ivanov, J.W. Head and A.S. Krasilnikov.
\newblock Geologic history of the south circumpolar region (SCR) of the Moon.
\newblock In {\em Icarus, Volume 394, 115422, ISSN 0019-1035}, 2023.
\newblock doi: \url{https://doi.org/10.1016/j.icarus.2022.115422}

\bibitem{Mazarico2010Illumination}
E. Mazarico, G.A. Neumann, D.E. Smith, M.T. Zuber and M.H. Torrence.
\newblock Illumination conditions of the lunar polar regions using LOLA topography
\newblock In {Icarus, Volume 211, 2, pages 1066--1081}, 2011.
\newblock doi: \url{https://doi.org/10.1016/j.icarus.2010.10.030}

\bibitem{Izzo2022Neuromorphic}
Dario Izzo, Alexander Hadjiivanov, Dominik Dold, Gabriele Meoni and Emmanuel Blazquez.
\newblock Neuromorphic computing and sensing in space.
\newblock In {Artificial Intelligence for Space: AI4SPACE, pages 107-159}, 2022.

\bibitem{Mahadevkar2022Review}
Supriya V. Mahadevkar, Bharti Khemani, Shruti Patil, Ketan Kotecha, Deepali R. Vora, Ajith Abraham and Lubna Abdelkareim Gabralla.
\newblock A Review on Machine Learning Styles in Computer Vision—Techniques and Future Directions.
\newblock In {IEEE Access, vol. 10, pp. 107293-107329}, 2022.
\newblock doi: \url{https://doi.org/10.1109/ACCESS.2022.3209825}

\bibitem{Pessia2019Artificial}
Romain Pessia, Genya Ishigami and Quentin Jodelet.
\newblock Artificial Lunar Landscape Dataset.
\newblock Kaggle, 2019.
\newblock doi: \url{https://doi.org/10.34740/kaggle/dsv/489236}

\bibitem{Prophesee}
Prophesee Metavision SDK Docs, 
\newblock \url{https://docs.prophesee.ai/stable/data/encoding_formats/evt3.html}
\end{thebibliography}

\end{document}